\documentclass{article}

\usepackage[final, nonatbib]{airov24}

\usepackage[utf8]{inputenc} 
\usepackage[T1]{fontenc}    
\usepackage[]{hyperref}       
\usepackage{url}            
\usepackage{booktabs}       
\usepackage{amsfonts}       
\usepackage{nicefrac}       
\usepackage{microtype}      
\usepackage{xcolor}         
\usepackage{graphicx}
\usepackage{subcaption}
\usepackage{float}
\usepackage{bm}
\usepackage{epsfig}
\usepackage{amsmath}
\usepackage{mathtools}
\usepackage[nameinlink,capitalize]{cleveref}
\usepackage{siunitx} 
\usepackage[style=ieee, backend=biber, hyperref=auto, maxbibnames=3]{biblatex}
\newcommand{\tp}{{\text{\scriptsize\sffamily T}}}
\title{Multimodal Medical Disease Classification with LLaMA~II}

\author{%
	Christian Gapp\textsuperscript{(1), (2)}\\
	\And
	Elias Tappeiner\textsuperscript{(1)}\\
	\And
	Martin Welk\textsuperscript{(1)}\\
	\And
	Rainer Schubert\textsuperscript{(1)}\\
	\And
	\vspace{-11pt}\\
	(1) Institute of Biomedical Image Analysis\\
	UMIT TIROL -- Private University for Health Sciences and Health Technology,\\	Eduard-Wallnöfer-Zentrum 1, 6060 Hall in Tirol, Austria\\
	\And
	\vspace{-11pt}\\
	(2) VASCage -- Centre on Clinical Stroke Research\\
	Innsbruck, Austria
	\And	
	\vspace{-11pt}\\
	\texttt{$\lbrace$christian.gapp, elias.tappeiner, martin.welk, rainer.schubert$\rbrace$}\\
	\texttt{@umit-tirol.at}\\
}
\begin{document}
	\maketitle
\begin{abstract}
Medical patient data is always multimodal. Images, text, age, gender, histopathological data are only few examples for different modalities in this context.
Processing and integrating this multimodal data with deep learning based methods is of utmost interest due to its huge potential for medical procedure such as diagnosis and patient treatment planning.
In this work we retrain a multimodal transformer-based model for disease classification. To this end we use the text-image pair dataset from OpenI consisting of 2D chest X-rays associated with clinical reports. Our focus is on fusion methods for merging text and vision information extracted from medical datasets. Different architecture structures with a LLaMA~II backbone model are tested.
Early fusion of modality specific features creates better results with the best model reaching $\approx$~97.10\% mean AUC than late fusion from a deeper level of the architecture (best model: $\approx$~96.67\% mean AUC). Both outperform former classification models tested on the same multimodal dataset.
The newly introduced multimodal architecture can be applied to other multimodal datasets with little effort and can be easily adapted for further research, especially, but not limited to, the field of medical AI.
\end{abstract}
\section{Introduction}
Deep learning from multimodal data, i.e. two or more modalities such as images, text, age, gender, histopathological data, becomes more and more interesting in medical diagnosis and can have a positive impact on patient treatment planning \cite{DeutscherRundfunk23}. We train a transformer-based model for disease classification, using a text-image pair dataset from OpenI \cite{OpenI_dataset, OpenI_dataset_creator} consisting of 2D chest X-rays combined with clinical reports.

The multimodal dataset was formerly processed in \cite{TransCheX2} for automated image annotation. With TransCheX \cite{TransCheX} -- published in MONAI (Medical Open Network AI) -- a transformer-based network for disease classification was trained on the same data \cite{OpenI_dataset}. For TransCheX BERT (Bidirectional Encoder Representations from Transformers) \cite{BERT} is used as backbone language model. 
The Transformer architecture was first introduced in \cite{AttentionIsAll} for text processing. With Vision Transformer \cite{16x16WORDS}, networks like CNNs for image processing (segmentation, classification, etc.) were partly outperformed. Nowadays transformer-based large language models are the state of the art in most deep learning based applications.
For this work, we use LLaMA~II \cite{LLaMAII} as backbone language model. LLaMA~II is one of the most recent large language models released by Meta~AI and can be freely used for research purposes. The model is available in the sizes 7B (7 billion model parameters), 30B and 70B. As the 7B model is already approximately 70 times larger than the BERT 100M parameter model (bert-base-uncased) used in TransCheX, all models trained here, instead of BERT, work with the 7B version of LLaMA~II.

Our architectures consist of three transformer-based parts, one for the text, one for the vision modality and one for processing both modalities at once. We focus on the latter part, i.e. the multimodal part. 
The survey Multimodal Learning with Transformers \cite{MultimodalTransformerFusion} summarizes fusion strategies for merging multimodal data. Our focus is on a method, that has cross layers in the architecture to fuse text and vision.
We introduce early, late and mixed fusion strategies by varying the position of the cross layer, i.e. the layer that merges the multimodal information. During training the model is fine-tuned to our specific task using LoRA \cite{LoRA} (Low Rank Adaptation) as a parameter efficient fine tuning method. Modifying several options of LoRA we train seven models for disease classification. 

The performance of the fine-tuned models is measured by their mean AUC (area under the ROC (receiver operating characteristic) curve) on the test dataset. Overall results are promising and highlight the potential of multimodal classification for medical diagnosis using large language model based architectures.

\section{Multimodal Medical Chest X-Ray Dataset}

\begin{figure}[]
	\begin{subfigure}[c]{0.32\textwidth}
		\includegraphics[width=\textwidth]{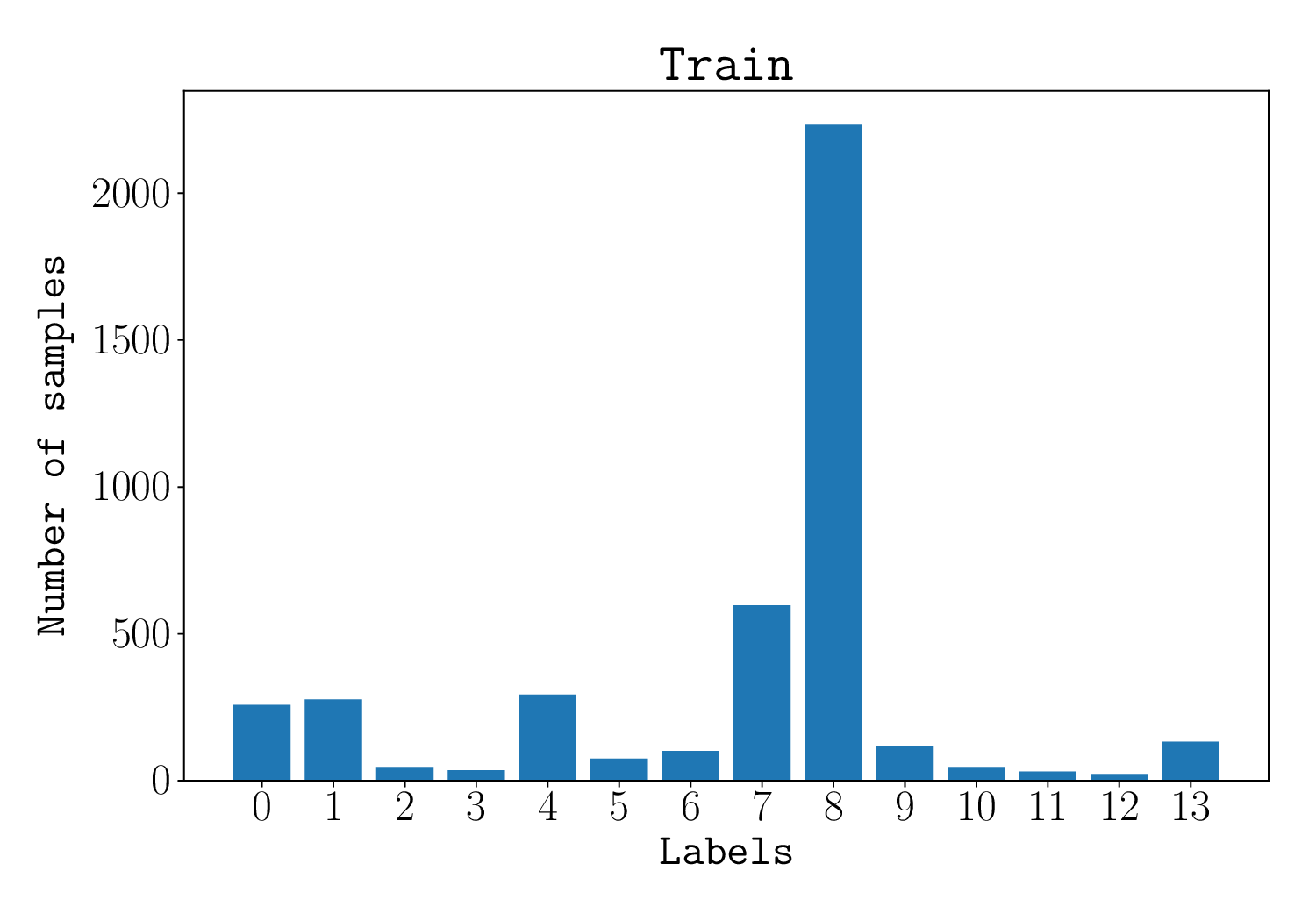}
		\centering
	\end{subfigure}
	\begin{subfigure}[c]{0.32\textwidth}
		\includegraphics[width=\textwidth]{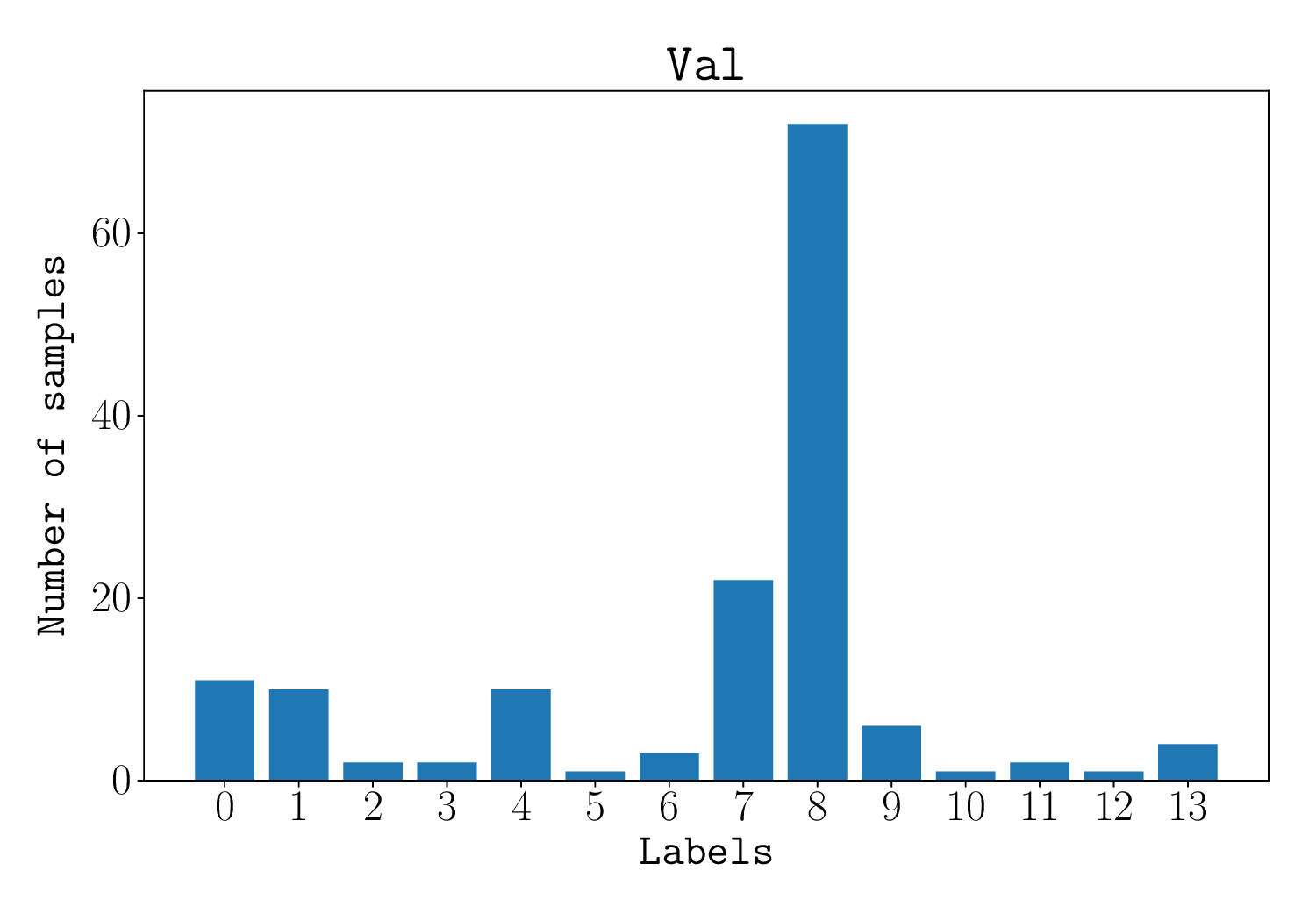}
		\centering
	\end{subfigure}
	\begin{subfigure}[c]{0.32\textwidth}
		\includegraphics[width=\textwidth]{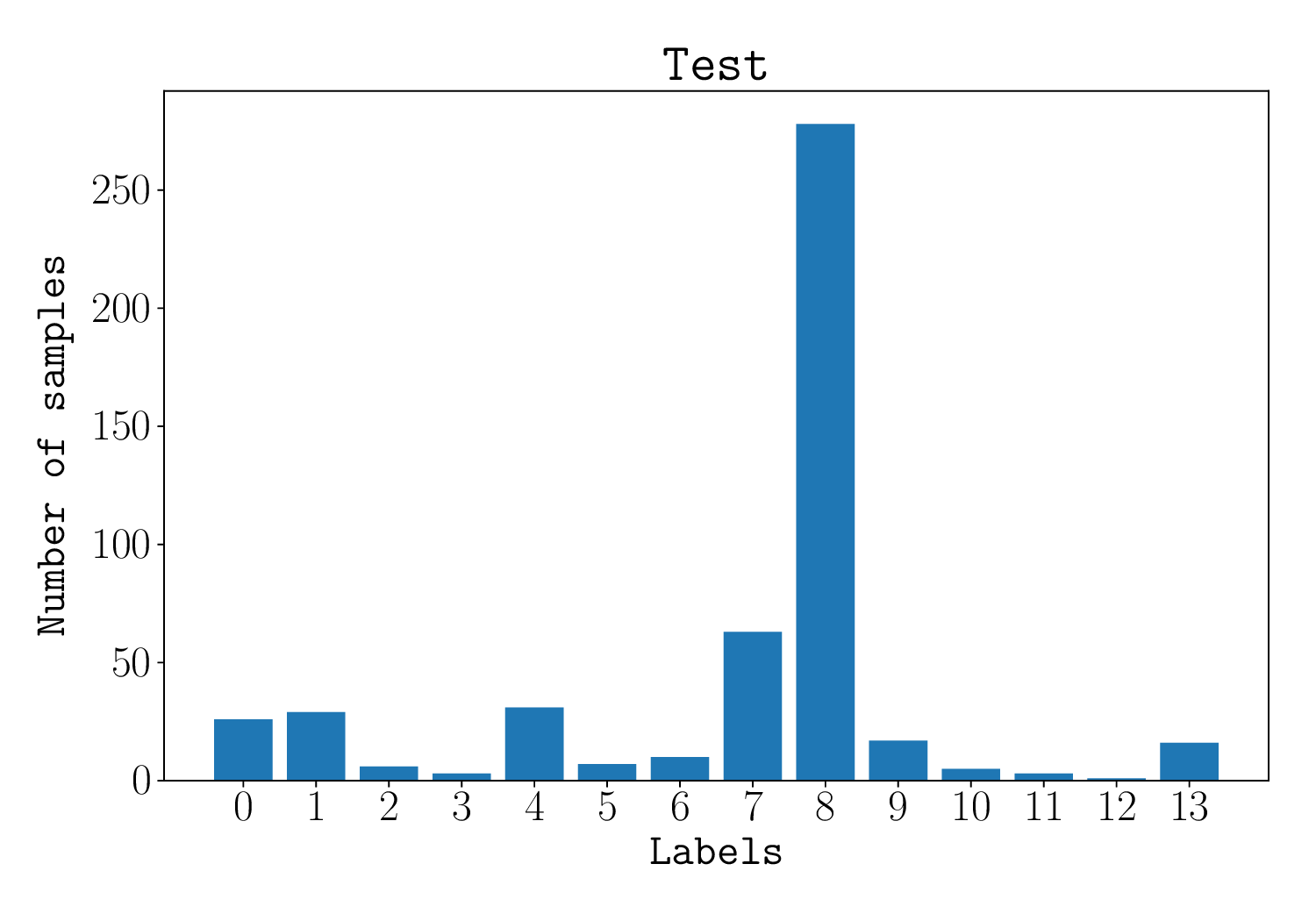}
		\centering
	\end{subfigure}
	\centering
	\caption{Distribution of labels in training, validation, and test datasets. Labels 0 to 13 represent the 14 classes: 0:~Atelectasis, 1:~Cardiomegaly, 2:~Consolidation, 3:~Edema, 4:~Enlarged-Cardiomediastinum, 5:~Fracture, 6:~Lung-Lesion, 7:~Lung-Opacity, 9:~Pleural-Effusion, 10:~Pleural Other, 11:~Pneumonia, 12:~Pneumothorax, 13:~Support-Devices, 8:~No Finding.}
	\label{fig-DataDistribution}
\end{figure}

The multimodal dataset used in our work is released by OpenI \cite{OpenI_dataset, OpenI_dataset_creator}. The dataset consists of 2D chest X-rays (256$\times$256), a clinical report and target classes (diseases, findings) for each patient. Training (3199 image-text pairs), validation (101) and test datasplits (377) are used equivalently as in \cite{TransCheX}. The distributions in their classification labels 0 to 13 (=~14 classes) are depicted in \cref{fig-DataDistribution}. Twelve labels depict diseases (0, 1, 2, 3, 4, 5, 6, 7, 9, 10, 11, 12), one is for Support-Devices (13) and one for No Finding (8).

\begin{figure}[]
	\centering
	\begin{minipage}[t]{.4\textwidth}
		\begin{figure}[H]
			\includegraphics[width=0.9\textwidth]{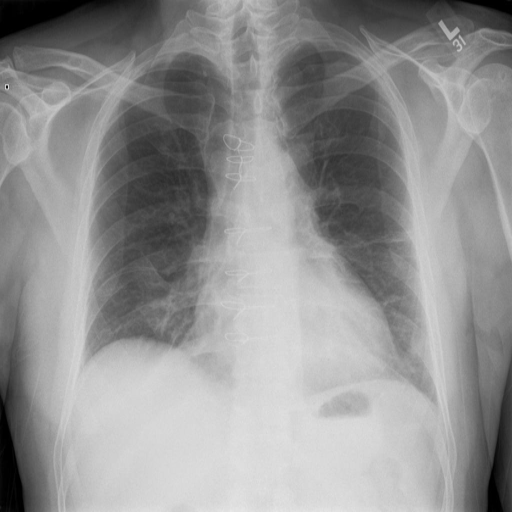}
		\end{figure}
	\end{minipage}
	\begin{minipage}[t]{.6\textwidth}
	\begin{figure}[H]
	\fbox{\parbox{\textwidth-6pt}{
		\texttt{FINDINGS : There are postoperative changes of sternotomy and CABG. There is stable mild cardiomegaly. There are scattered XXXX of subsegmental atelectasis, decreased from the prior chest radiograph. No focal airspace consolidation. No pleural effusion or pneumothorax. There are minimal degenerative changes of the spine.\newline
		IMPRESSION : 1. Scattered bilateral subsegmental atelectasis. Decreased from prior radiograph. 2. Stable mild cardiomegaly.
		}
	}}
	\end{figure}
	\end{minipage}
	\caption{Example training item CXR1053\_IM-0040-1001. Diseases: Atelectasis, Cardiomegaly, Enlarged Cardiomediastinum, Lung Opacity.}%
	\label{fig:example}
\end{figure}

As one patient can have several diseases, multiple classes are possible, except there was No Finding. \cref{fig:example} shows an example 2D chest X-ray with the associated clinical report.

\section{Model -- Architecture}
When processing multimodal data the fusion of two or more modalities is a central part. Different fusion methods involve different architectures. Our main focus is on the cross layer, i.e. the layer that merges the two modalities.
In order to demonstrate the different fusion strategies and their architectures, we first outline the details of the feature extraction from text and vision separately. 

\subsection{Feature Extraction}
The multimodal data in its original form can not be analyzed from a deep learning model. Therefore, at first both clinical report (text) and 2D chest X-rays (vision) must be preprocessed before being forwarded to the models input. The two different modalities are thus transformed into the same representative space.

\paragraph{Text Features}
The text tokenizer used for the LLaMA~II \cite{LLaMAII} language model first splits the clinical report into syllables and then encodes them into a vector of numbers. The final text features for the model input are the embeddings of the encoded text syllables with an additional positional embedding. For each clinical report 4096 features are generated.

In \cref{tab-textfeatureExtraction} we give an example extraction process from input text to text features.
\begin{table}[]
	\caption{Text Feature Extraction. Example. The input text is first split into tokenizer syllables, then transformed into tokenizer numbers and with additional positional information finally concatenated to text features.}
	\centering
	\begin{tabular}{ll}
		\toprule
		\textbf{input text:} & FINDINGS : The cardiomediastinal silhouette \dots \\[5.5pt] 
		\textbf{tokenizer syllables:} & [['F', 'IND', 'ING', 'S'], [':'], ['The'], ['card', 'iom', 'ed',\\
		& 'iast', 'inal'], ['sil', 'hou', 'ette'] \dots]\\[5.5pt]
		\textbf{tokenizer numbers:} & [[383, 22255, 4214, 29903], [584], [450], [5881, 14910, 287,\\
		& 15736, 979], [4047, 10774, 2353] \dots]\\[5.5pt]
		\textbf{text features} & [$t_0, t_1, \dots, t_{4093}$], with $t \in \mathbb{R}$  \\
		\bottomrule
	\end{tabular}
	\label{tab-textfeatureExtraction}
\end{table}

\paragraph{Vision Features}
With 2D convolution we extracted vision features from the 2D chest X-rays.
The method we used applies a convolution over an input signal (here: image) composed of several input planes.
First, we split the image (256$\times$256) into 16 patches of size 32$\times$32 pixels. With a linear, learnable transformation, 2D convolution, finally 4096 vision features per image corresponding to the embedding dimension used for text feature extraction are produced.

\subsection{Fusion Strategies}

We want to introduce different fusion strategies for merging the two modalities, text and vision.
All of our strategies have cross layers for multimodal data fusion in common. 
Regarding the location of these cross layers, early, late or mixed fusion is possible.

\subsubsection{Cross Layer: Multimodal Fusion}
Text and vision layers process one modality each. With cross layers, the modalities are fused together with the purpose of exchanging information from the data. Text, and vision and cross layers consist of attention blocks that compute 
\begin{equation}\label{eq:attention}
	\mathrm{Attention}(\bm{Q},\bm{K},\bm{V}) = \mathrm{softmax} \left( \frac{\bm{Q}\bm{K}^\tp}{\sqrt{d_k}}\right)\bm{V} \text{,}
\end{equation}
with matrix $\bm{Q}$ containing the queries, $\bm{K}$ containing the keys and $\bm{V}$ the values. $\sqrt{d_k}$ is a scaling factor, with $d_k$ representing the dimension of keys. Multi-head attention is realized with an additional linear layer. Multiple attention-heads produced with \cref{eq:attention} are concatenated and projected using another linear layer $\bm{W}_O$. With 
\begin{equation}\label{eq:multi_head}
	\mathrm{multi\_head\_attention}(\bm{Q}, \bm{K}, \bm{V}) = \mathrm{Concat}(\mathrm{head_1}, \dots, \mathrm{head}_h)\bm{W}_O \text{,}
\end{equation}
where \mbox{$\mathrm{head_i} = \mathrm{Attention}(\bm{Q}, \bm{K}, \bm{V})_i$} with \mbox{$i = 1, \dots, h$} (number of heads), 
the model is capable of attending to different information in parallel \cite{AttentionIsAll}.

Text, vision and cross layers further contain a feed forward module 
\begin{equation}\label{eq:feed-forward}
	\mathrm{feed\_forward}(x) = \bm{W}_2(\mathrm{silu}(\bm{W}_1(x)\bm{W}_3(x))) \text{,}
\end{equation}
with $\bm{W}_1$, $\bm{W}_2$, $\bm{W}_3$ representing linear layers and silu$(x) = x \sigma(x)$, being the Sigmoid Linear Unit function with the logistic sigmoid $\sigma(x)$.
Taking the attention block's output
\begin{equation}\label{eq:h}
	y = x + \mathrm{multi\_head\_attention}({L}^2\mathrm{norm}(x))\text{,}
\end{equation}
with input $x$, the final text, vision layers' output is computed by
\begin{equation}\label{eq:out}
	\mathrm{out} = y + \mathrm{feed\_forward}({L}^2\mathrm{norm}(y)).
\end{equation}
The cross layer gets a query from one modality and a key-value pair from the other modality. Hence we need at least two cross layers (1: query (vision), key-value (text), 2: query (text), key-value (vision)) per level. This can be interpreted as asking for text (key, value) from vision (query) (1) and vice versa (2).

\subsubsection{Parallel Pipeline -- Early Fusion}
\begin{figure}[]
	\includegraphics[width=0.6\textwidth]{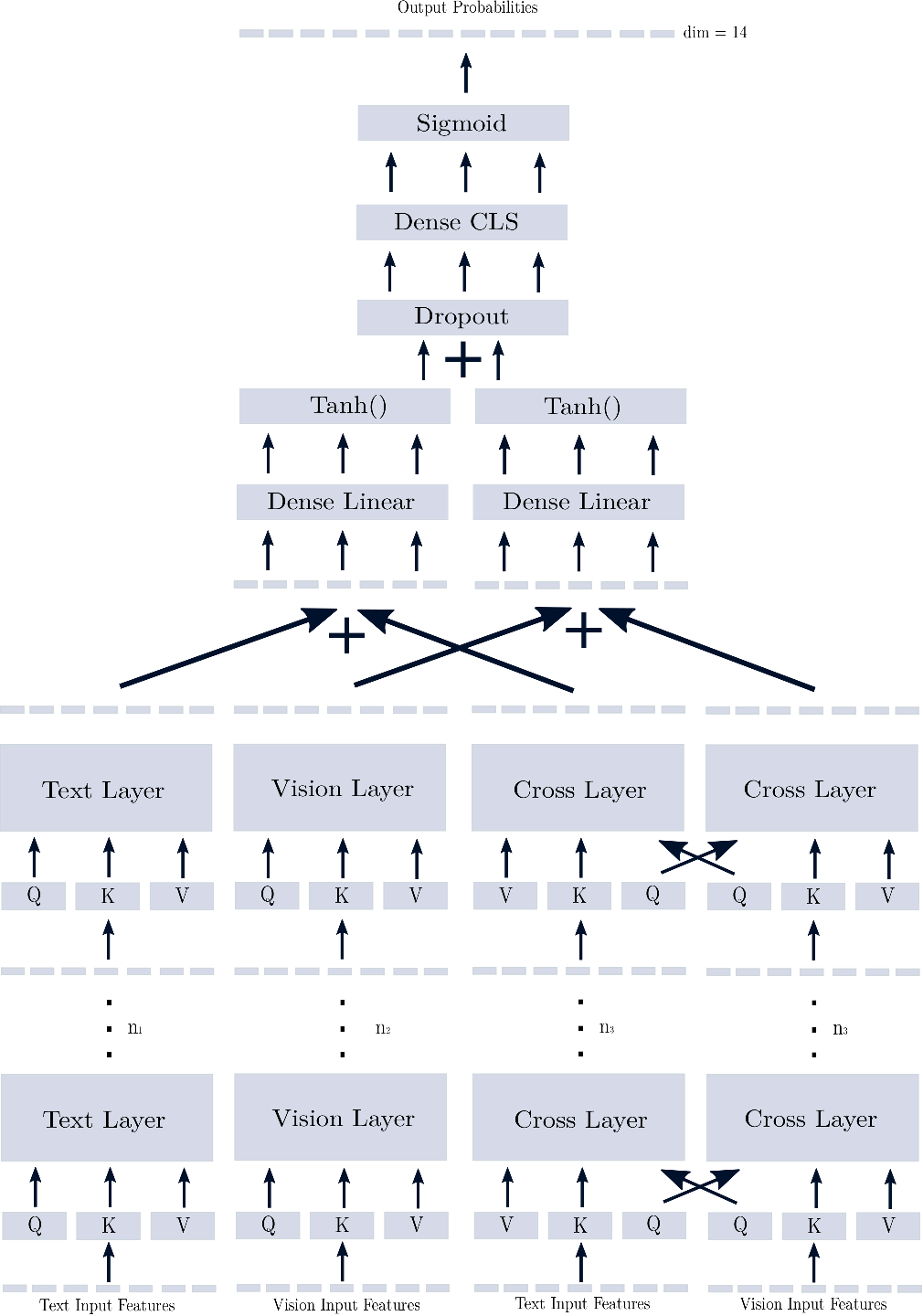}
	\centering
	\caption{Architecture: Parallel Workflow}
	\label{fig-ParallelArch}
\end{figure}
The parallel architecture is depicted in \cref{fig-ParallelArch}. The cross layers are parallel to the text and vision layers at each level.
Herein the untouched text and vision features are fed to the cross layers' input. As the fusion of the modalities is already done at the first level, parallel architectures generate an early fusion of multimodal information.

Dense Linear, Dense CLS (see \cref{fig-ParallelArch}) are linear layers with different input and output dimensions to finally get the output size down to 14 (number of classes). Tanh() transforms the Dense Linears' output into [-1, 1]. The Dropout layer realizes a regularization method, that drops some layer outputs randomly according to a specific rate (here: 0.1) \cite{Dropout}. The Sigmoid activation function generates output probabilities individual for each class.

\subsubsection{Serial Pipeline -- Late Fusion}
\begin{figure}[]
	\includegraphics[width=0.7\textwidth]{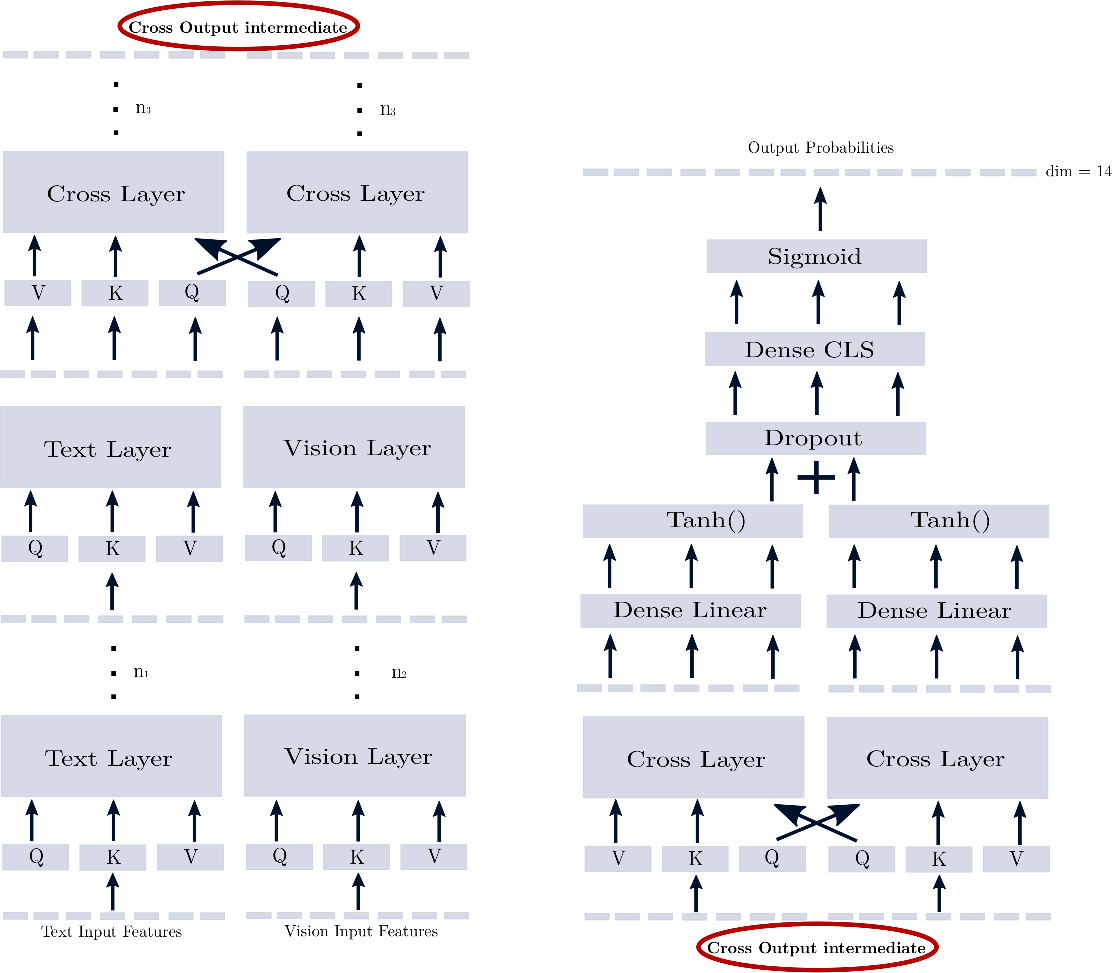}
	\centering
	\caption{Architecture: Serial Workflow}
	\label{fig-SerialArch}
\end{figure}
With the serial architecture, see \cref{fig-SerialArch}, the cross layer is put after the two modality specific layers (text, vision layer). Thereby the processed text and vision features are fed to the cross layers instead of using the untouched ones as done in the parallel workflow. Late Fusion combines the multimodal information after the last level of single modality specific layers.

The pipeline after the cross layers is equal to the parallel one as described above.

\subsubsection{Mixed Pipeline -- Mixed Fusion}
As a last scenario a mixed version of the parallel and serial pipeline is trained. Starting from the parallel architecture a cross layer is added immediately before the Dense Linear layers. In order to let the size of parameters unchanged one cross layer was removed from the former parallel architecture. In other words, one parallel cross layer is replaced by a serial one.

\section{Training Methodologies}
The enormous size of large language models generally leads to difficulties during training, as they require a lot of GPU memory. Therefore, we introduce a parameter efficient fine tuning method that reduces GPU memory requirement and thus makes it possible to train architectures with a huge number of model parameters.

\subsection{Low Rank Adaptation (LoRA)}
LoRA \cite{LoRA} enables to train a small amount of model parameters by freezing the pretrained model weights and injecting trainable rank decomposition matrices. This reduces both gpu memory requirement and computation time whilst maintaining the same or resulting in an even better performance. Let \mbox{$\bm{W} \in \mathbb{R}^{d\times k}$} contain the weights for one matrix in the layer $L$, then $\bm{A} \in \mathbb{R}^{r\times k} $ and $\bm{B} \in \mathbb{R}^{d\times r}$ represent the two additional LoRA matrices for this matrix $\bm{W}$ in $L$. 
Instead of training all \mbox{$d\times k$} parameters, now only the \mbox{$r \times k$} parameters from $\bm{A}$ and the \mbox{$d\times r$} parameters from $\bm{B}$ are updated during training routine and added to the frozen matrix $\bm{W}$. The weights in $\bm{B}$ are initialized with zeros, whereas Random Gaussian initialization is applied for $\bm{A}$. For input $x$, formerly the corresponding output was $h = \bm{W}x$. Now we have
\begin{equation}
	h = \bm{W}_0x + \Delta \bm{W}x = \bm{W_0}x + \bm{BA}x \text{,}
\end{equation}
where $\bm{W}_0$ is the frozen weight matrix at initial state.
The hyperparameter \mbox{$r \ll d, k $} directly impacts the parameter reduction. Hereby memory requirement is enormously reduced by factor $r(1/d+1/k)$ for each lora-layer. Additional LoRA parameters are the \mbox{lora\_dropoutrate $\in \mathbb{R} \cap [0,1]$} and scaling factor \mbox{$\alpha \in \mathbb{N}$}. We set them constant as mentioned in the Ablation Studies afterwards.

In our implementation we use a model wrapper from Parameter Efficient Fine Tuning (PEFT) \cite{PEFT} in order to handle the LoRA layers.

\cref{tab-targetmoduleslora} outlines the layers (target modules) we apply within LoRA for all architectures. $\bm{W}_Q$, $\bm{W}_K$, $\bm{W}_V$ represent the weight matrices for $Q$, $K$, $V$ respectively. $\bm{W}_O$ is the projection matrix from  \cref{eq:multi_head}. $\bm{W}_1$, $\bm{W}_2$, $\bm{W}_3$ are from the feed forward module used in \cref{eq:feed-forward}.
\begin{table}[]
	\caption{Target Modules for LoRA}
	\centering
	\begin{tabular}{lllll}
		\toprule
		\multicolumn{2}{c}{LoRA Layer} & \multicolumn{3}{c}{Modality}         \\
		\cmidrule(r){1-2}
		\cmidrule(l){3-5}
		Name     & Type    & Vision & Text & Cross\\
		\midrule
		$\bm{W}_Q$    	& Linear Layer & \multicolumn{1}{c}{ \checkmark } & \multicolumn{1}{c}{ \checkmark } & \multicolumn{1}{c}{ \checkmark }    \\
		$\bm{W}_K$    	& Linear Layer & \multicolumn{1}{c}{ \checkmark } & \multicolumn{1}{c}{ \checkmark } & \multicolumn{1}{c}{ \checkmark }\\
		$\bm{W}_V$    	& Linear Layer & \multicolumn{1}{c}{ \checkmark } & \multicolumn{1}{c}{ \checkmark } & \multicolumn{1}{c}{ \checkmark }\\
		$\bm{W}_O$		& Linear Layer & \multicolumn{1}{c}{ \checkmark } & \multicolumn{1}{c}{ \checkmark } & \multicolumn{1}{c}{ \checkmark }\\
		$\bm{W}_1$		& Linear Layer & \multicolumn{1}{c}{ - } & \multicolumn{1}{c}{ - } & \multicolumn{1}{c}{ - }   \\
		$\bm{W}_2$		& Linear Layer & \multicolumn{1}{c}{ \checkmark } & \multicolumn{1}{c}{ \checkmark } & \multicolumn{1}{c}{ \checkmark }\\
		$\bm{W}_3$		& Linear Layer & \multicolumn{1}{c}{ - } & \multicolumn{1}{c}{ - } & \multicolumn{1}{c}{ - }   \\
		embeddings  & Embedding  & \multicolumn{1}{c}{ - } & \multicolumn{1}{c}{ \checkmark } & \multicolumn{1}{c}{ - }\\
		vision\_proj     & 2D Conv & \multicolumn{1}{c}{ \checkmark } & \multicolumn{1}{c}{ - } & \multicolumn{1}{c}{ - }\\
		\midrule
		Dense Linear & Linear Layer & \multicolumn{3}{c}{ $-$ }  \\
		Dense CLS & Linear Layer & \multicolumn{3}{c}{ $-$ } \\
		\bottomrule
	\end{tabular}
	\label{tab-targetmoduleslora}
\end{table}

\subsection{Ablation Studies}
All architectures are of the same size regarding number of parameters. They contain three text, three vision and three cross layers. Note that for the mixed architecture we positioned two cross layers parallel to the text and vision layers, and one cross layer serial to all of them. The feature extraction (input generation) and the cross layers' output processing -- Dense Linear $\rightarrow$ Tanh() $\rightarrow$ Dropout $\rightarrow$ Dense CLS $\rightarrow$ Sigmoid -- to finally compute the output probabilities remains unchanged in all scenarios.

By varying the position of the cross layers seven models (three for serial, three for parallel, one for mixed pipeline) are trained, tested on the test datasplit and compared using the AUC (area under the ROC (receiver operating characteristic) curve) metric. The LoRA parameter $r$ is modified for the serial and parallel pipelines with \mbox{$r \in \lbrace 2,4,8 \rbrace$}, resulting in three models each. Additionally a mixed version with \mbox{$r$ = 2} is trained.
The other LoRA parameters are set as follows: \mbox{lora\_dropoutrate = 0.1}, \mbox{$\alpha$ = 32}.

\subsection{Training Details}

We implemented our models in python with the use of Pytorch \cite{PYTORCH}. The models were trained on a Nvidia Titan RTX 24 GB GPU with 15 epochs and \mbox{batch\_size = 20}. For optimization the Adam Optimizer with \mbox{learning\_rate = $\num{e-4}$} and \mbox{weight\_decay = $\num{e-5}$} was used.
We applied binary cross entropy as loss function for our multi-class classification problem that outputs probabilities~$\in$~[0,1] for each class.

%
%
%
\section{Results}
TransCheX's \cite{TransCheX} model's evaluation statistics on the test dataset from OpenI \cite{OpenI_dataset, OpenI_dataset_creator} resulted in $96.29\%$ 
mean AUC. 
Our results for parallel, serial and mixed architectures will be presented below.
\begin{table}[]	
	\caption{Performance (AUC) of the three architectures (serial, parallel, mixed) for different LoRA parameters $r$ on the test dataset. The AUC is printed for each class and each training set. In the final row the arithmetic mean is computed column by column. The last column contains the results from TransCheX \cite{TransCheX}.}
	\centering´
	\begin{tabular}{lcccccccc}
		\toprule
		& \multicolumn{3}{c}{parallel} & \multicolumn{3}{c}{serial}  & \multicolumn{1}{c}{mixed} &  \\
		\cmidrule(r){2-4}
		\cmidrule(r){5-7}  
		\cmidrule(r){8-8}
		Class     & $r=2$ & $r=4$ & $r=8$ & $r=2$ & $r=4$ & $r=8$ & $r=2$ & TransCheX \\
		\midrule
		Atelectasis &$0.984$ &$0.970$ &${0.986}$ &$0.984$ &$0.986$ &$0.986$ & ${0.987}$ & $0.993$\\
		Cardiomegaly &${0.999}$ &$0.995$ &$0.999$ &$0.997$ &$0.994$ &${0.998}$ & $0.987$ & $0.975$ \\
		Consolidation &$0.928$ &${0.934}$ &$0.931$ &$0.971$ &$0.960$ &${0.974}$ & $0.944$ &$0.953$\\
		Edema     &${0.998}$ &${0.998}$ &$0.996$ &$0.996$ &$0.994$ &${0.998}$ & ${0.999}$ &$0.990$\\
		Enlarged-Card. &${0.997}$ &$0.994$ &$0.996$ &$0.994$ &${0.989}$ &$0.995$ & $0.984$ &$0.945$\\
		Fracture &$0.989$ &$0.983$ &${0.992}$ &$0.982$ &$0.975$ &${0.987}$ & $0.980$ &$0.991$\\
		Lung-Lesion &$0.974$ &${0.976}$ &$0.976$ &$0.958$ &${0.977}$ &$0.962$ & $0.966$ &$0.947$\\
		Lung-Opacity   &$0.979$ &$0.984$ &${0.984}$ &$0.983$ &${0.984}$ &$0.982$ & $ 0.984$ &$0.986$\\
		No-Finding &${0.975}$ &$0.968$ &$0.968$ &$0.968$ &$0.971$ &${0.971}$ & $0.968$ &$0.957$\\
		Pleural-Effusion &${0.917}$ &$0.905$ &$0.909$ &${0.944}$ &$0.906$ &$0.918$ & $0.926$ &$0.898$\\
		Pleural\_Other &$0.963$ &$0.926$ &${0.987}$ &${0.951}$ &$0.875$ &$0.937$ & $0.993$ &$0.997$\\
		Pneumonia     &$0.986$ &$0.980$ &${0.989}$ &${0.995}$ &$0.988$ &$0.980$ & $0.964$ &$0.971$\\
		Pneumothorax &${0.973}$ &$0.955$ &$0.880$ &$0.856$ &${0.976}$ &$0.774$ & $0.840$ &$0.979$\\
		Support-Devices &$0.928$ &$0.941$ &${0.946}$ &$0.932$ &${0.959}$ &$0.925$ & $0.918$ &$0.898$\\
		\midrule
		Mean	& $\mathbf{0.971}$ & $0.969$ & $0.967$ & $0.965$ & $\mathbf{0.967}$ & $0.956$ & $0.960$ & $0.963$\\
		\bottomrule
	\end{tabular}
	\label{tab-performanceTestDataset_all}
\end{table}
\begin{figure}[]
	\begin{subfigure}[c]{0.3\textwidth}
		\includegraphics[width=\textwidth]{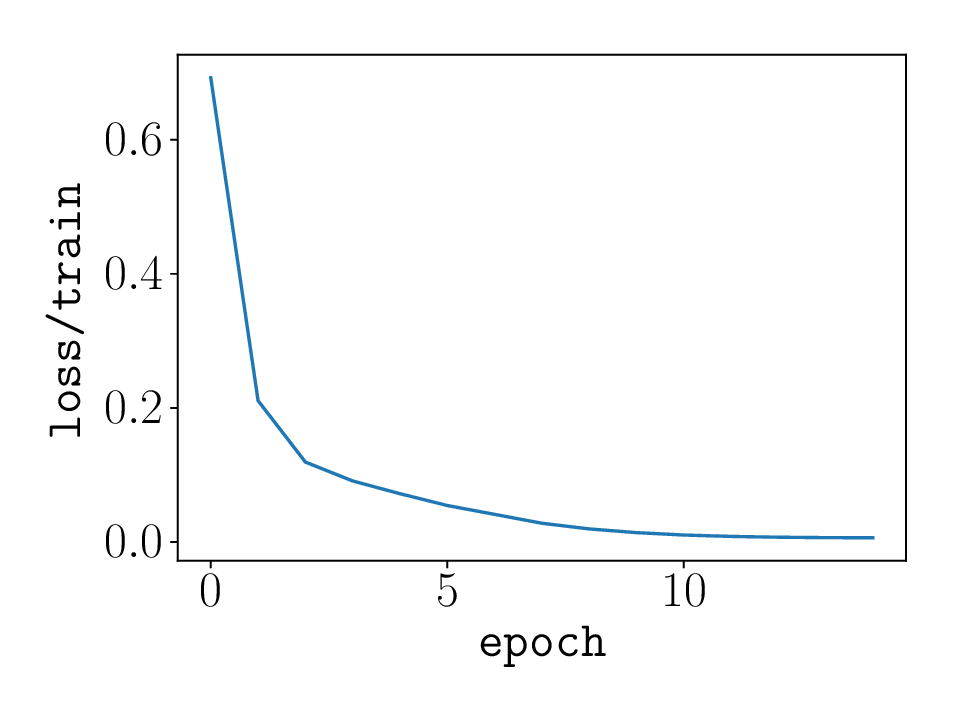}
		\centering
		\caption{train loss}
	\end{subfigure}
	\begin{subfigure}[c]{0.3\textwidth}
		\includegraphics[width=\textwidth]{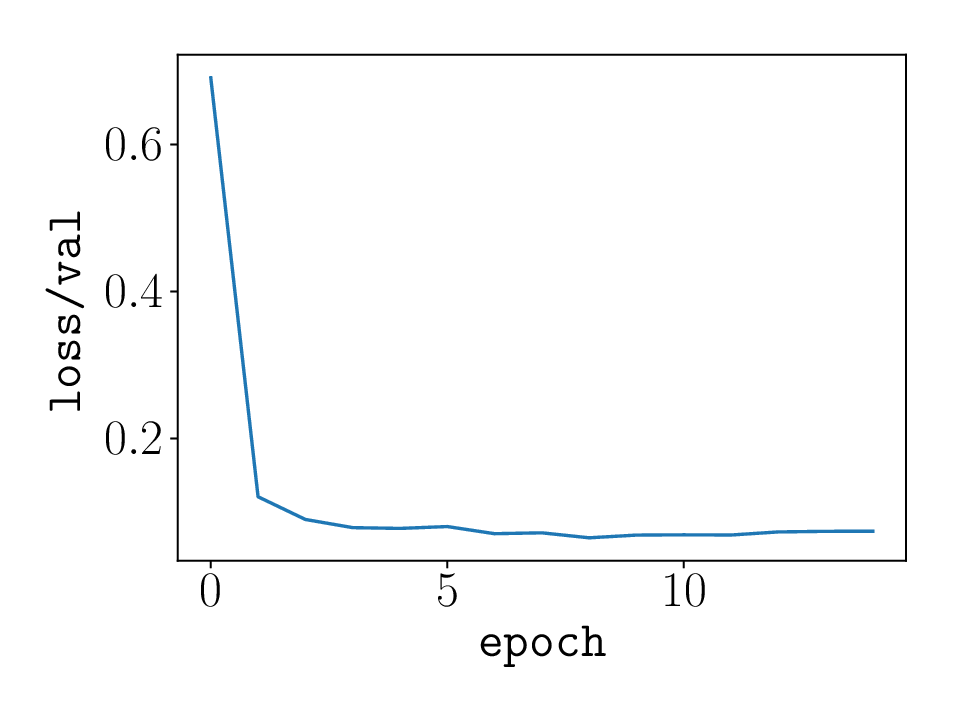}
		\centering
		\caption{validation loss}
	\end{subfigure}
	\begin{subfigure}[c]{0.3\textwidth}
		\includegraphics[width=\textwidth]{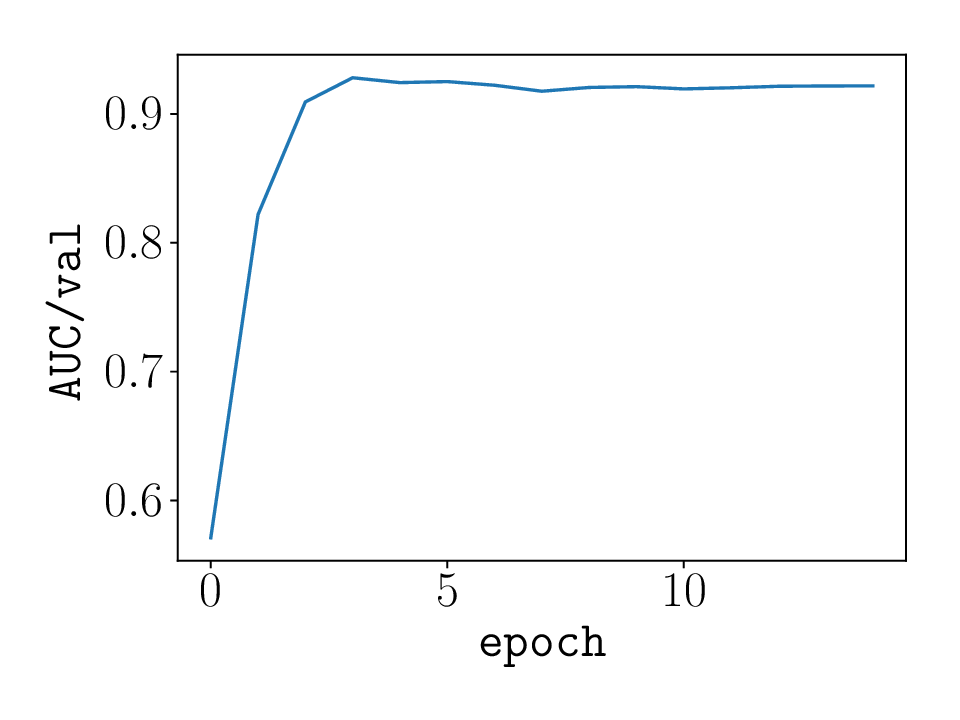}
		\centering
		\caption{validation AUC}
	\end{subfigure}
	\centering
	\caption{Metrics of training with parallel pipeline and LoRA parameter $r=2$. (a) and (b) show the losses on the training and validation dataset for epochs 0 to 14. With (c) the performance (AUC) on the validation dataset is depicted.}
	\label{fig:results_parallel_r2}
	\begin{subfigure}[c]{0.3\textwidth}
		\includegraphics[width=\textwidth]{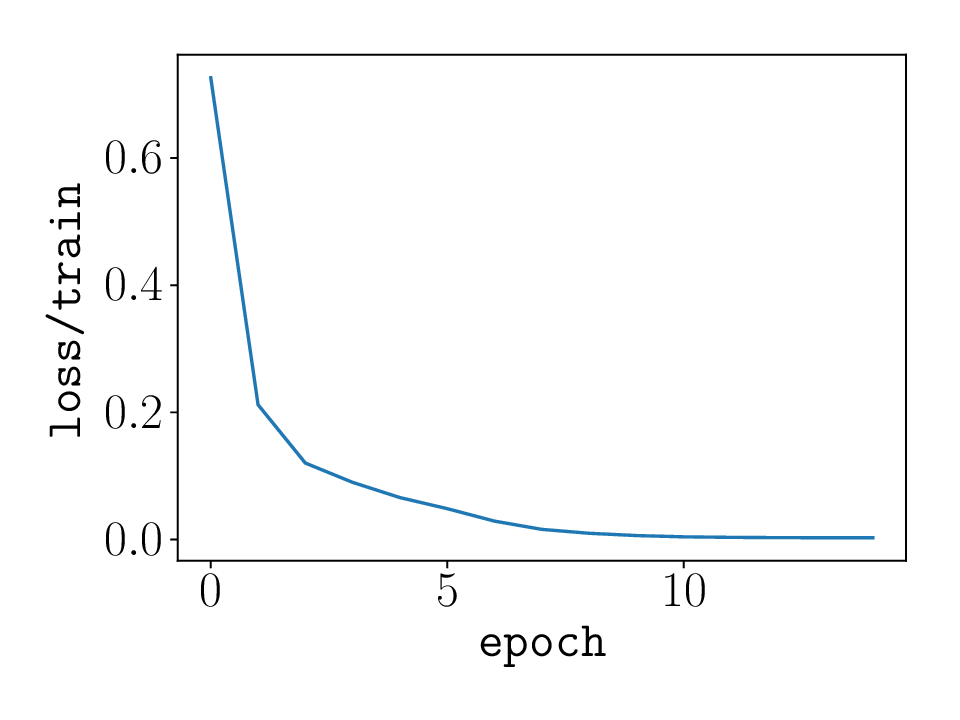}
		\centering
		\caption{train loss}
	\end{subfigure}
	\begin{subfigure}[c]{0.3\textwidth}
		\includegraphics[width=\textwidth]{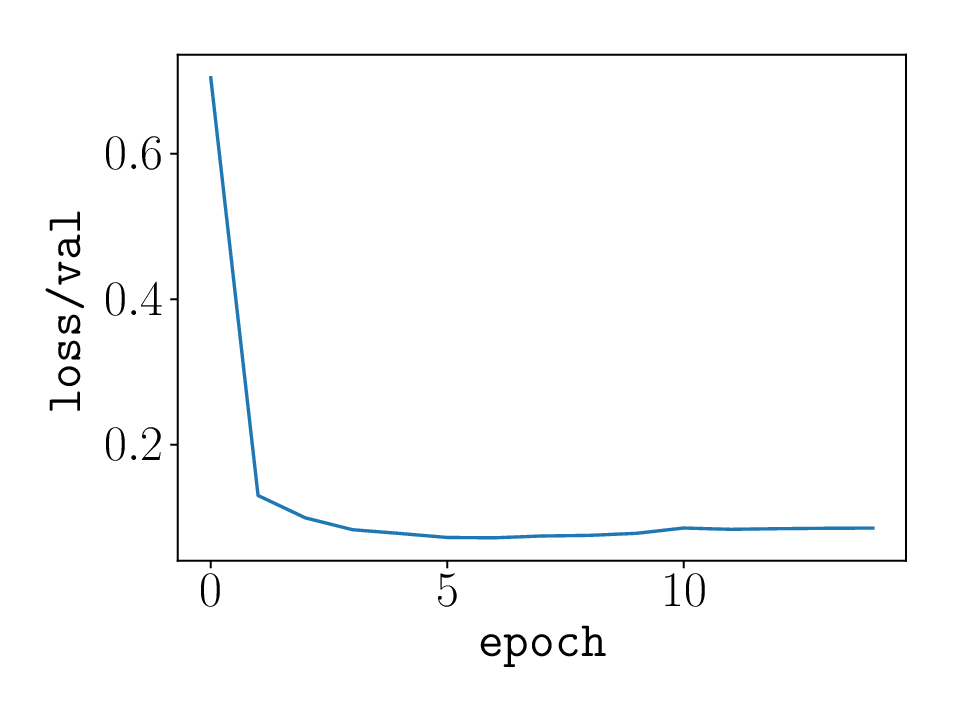}
		\centering
		\caption{validation loss}
	\end{subfigure}
	\begin{subfigure}[c]{0.3\textwidth}
		\includegraphics[width=\textwidth]{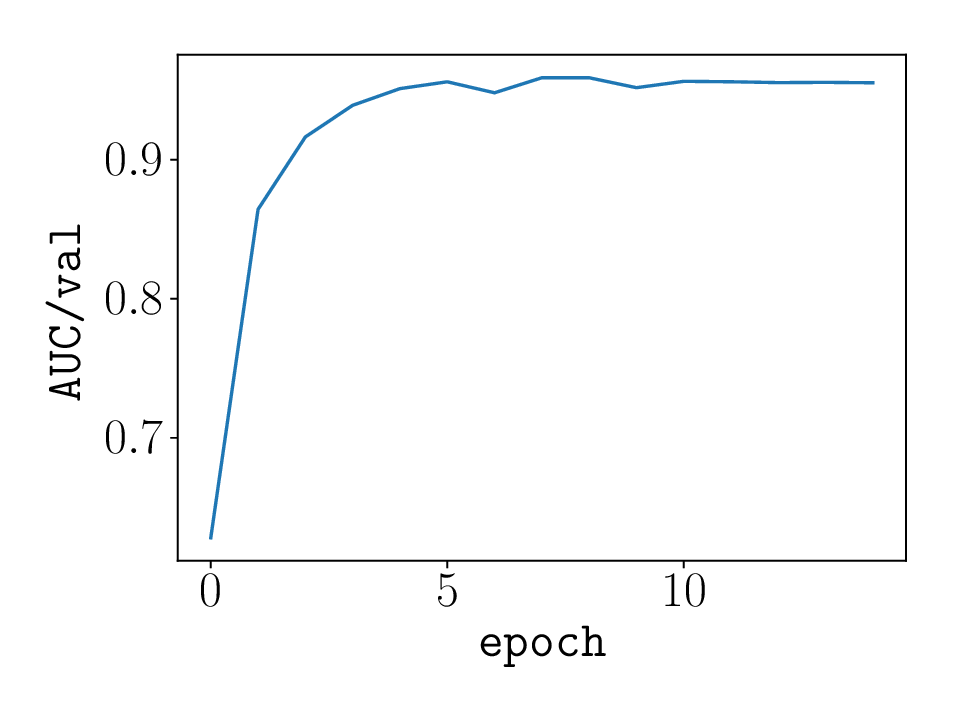}
		\centering
		\caption{validation AUC}
	\end{subfigure}
	\caption{Metrics of training with serial pipeline and LoRA parameter $r=4$. In (a) and (b) the losses on the training and validation dataset are depicted for each epoch 0 to 14. (c) shows the AUC metric on the validation dataset.}
	\label{fig:results_serial_r4}
\end{figure}

\paragraph{Parallel Pipeline}
The model with parallel pipeline and LoRA parameter $r$ = 2 has the best performance on the test datasplit with 97.10\% mean AUC and hence outperforms the TransCheX's model by $0.81 \%$, see \cref{tab-performanceTestDataset_all}. In \cref{fig:results_parallel_r2} training and validation details are depicted.

\paragraph{Serial Pipeline}
The model with LoRA parameter $r=4$ performs best with 96.67\% mean AUC on the test datasplit and thus slightly better than the TransCheX's model by $0.38\%$, see also \cref{tab-performanceTestDataset_all}. Training and validation details can be viewed in \cref{fig:results_serial_r4}.

\paragraph{Mixed Pipeline}
With a mean AUC = 96.00\%, 
the mixed pipeline performs marginally worse than the other architectures. The class specific AUC can be viewed in \cref{tab-performanceTestDataset_all} (second last column).

%
%
%

\section{Discussion}
\label{sec-discussion}
In our tests, five out of seven architectures -- all three parallel (LoRA: \mbox{$r \in \lbrace 2,4, 8\rbrace$)} and two serial ones (LoRA: \mbox{$r \in \lbrace 2,4\rbrace$)} -- outperformed former models tested on the dataset from OpenI with regard to the mean AUC. Solely the serial with \mbox{LoRA: $r$ = 8} and the mixed pipeline performed worse. All in all the parallel architecture with \mbox{LoRA: $r$ = 2} leads to the best result with mean \mbox{AUC = 97.10\%}, outperforming the current state of the art. Regarding the best parallel \mbox{(LoRA: $r$ = 2)} and the best serial pipeline \mbox{(LoRA: $r$ = 4)} the early fusion (parallel architecture) of the two modalities text (clinical reports) and vision (2D chest X-rays) seems to perform slightly better than the late fusion (serial architecture). In the case of the multimodal dataset processed here, this means, the early fusion of the raw, untouched features lead to better classification performance than processing the modality specific outputs from a deeper level as done with late fusion.

The replacement of BERT with the larger language model LLaMA~II~7B, fine tuning with LoRA, and the optimization of the position to merge multimodal information build the central part to gain such great performance for disease classification on the text-image pair dataset from OpenI. This can have remarkable impact on the future patient treatment planning in medicine \cite{DeutscherRundfunk23}.


\section{Summary and Outlook}
In this paper we trained a multimodal transformer-based model with a LLaMA~II backbone for medical disease classification. The underlying multimodal dataset from OpenI consists of 2D chest X-rays with corresponding clinical reports. We focused on different fusion strategies for merging the text with vision data. Herein the position of the cross layer, i.e. the layer that fuses the multimodal information, was central. Utilizing LoRA (Low Rank Adaptation) we trained seven models and tested them on the test datasplit from OpenI. Five out of seven models outperformed recent works (TransCheX \cite{TransCheX}) tested on the same dataset. Early fusion realized with the parallel architecture performed slightly better than late or mixed fusion. Our models can be applied to other multimodal datasets with little effort. Furthermore, our models build the foundation to study interpretability methods on multimodal data, which is a special interest in the medical domain.

\begin{ack}
This study is supported by VASCage – Centre on Clinical Stroke Research. VASCage is a COMET Centre within the Competence Centers for Excellent Technologies (COMET) programme and funded by the Federal Ministry for Climate Action, Environment, Energy, Mobility, Innovation and Technology, the Federal Ministry of Labour and Economy, and the federal states of Tyrol, Salzburg and Vienna. COMET is managed by the Austrian Research Promotion Agency (Österreichische Forschungsförderungsgesellschaft).
FFG Project number: 898252
\end{ack}

\printbibliography[title=References]

@INPROCEEDINGS {TransCheX2,
author = {H. Shin and K. Roberts and L. Lu and D. Demner-Fushman and J. Yao and R. M. Summers},
booktitle = {2016 IEEE Conference on Computer Vision and Pattern Recognition (CVPR)},
title = {Learning to Read Chest X-Rays: Recurrent Neural Cascade Model for Automated Image Annotation},
year = {2016},
volume = {},
issn = {1063-6919},
pages = {2497-2506},
keywords = {diseases;x-rays;context;training;radiology;biomedical imaging;recurrent neural networks},
%doi = {10.1109/CVPR.2016.274},
%url = {https://doi.ieeecomputersociety.org/10.1109/CVPR.2016.274},
publisher = {IEEE Computer Society},
address = {Los Alamitos, CA, USA},
month = {jun}
}

@ARTICLE {TransCheX,
author = {Ali Hatamizadeh}, %et al ...
title = {TransCheX: Self-Supervised Pretraining of Vision-Language Transformers for Chest X-ray Analysis},
note = {\url{https://github.com/Project-MONAI/tutorials/blob/main/multimodal/openi_multilabel_classification_transchex/transchex_openi_multilabel_classification.ipynb} (Accessed: 2023-19-03)}
%year = {}
}

@ARTICLE{LLaMAII,
      title={Llama 2: Open Foundation and Fine-Tuned Chat Models}, 
      author={Hugo Touvron and Louis Martin and Kevin Stone and Peter Albert and Amjad Almahairi and Yasmine Babaei and Nikolay Bashlykov and Soumya Batra and Prajjwal Bhargava and Shruti Bhosale and Dan Bikel and Lukas Blecher and Cristian Canton Ferrer and Moya Chen and Guillem Cucurull and David Esiobu and Jude Fernandes and Jeremy Fu and Wenyin Fu and Brian Fuller and Cynthia Gao and Vedanuj Goswami and Naman Goyal and Anthony Hartshorn and Saghar Hosseini and Rui Hou and Hakan Inan and Marcin Kardas and Viktor Kerkez and Madian Khabsa and Isabel Kloumann and Artem Korenev and Punit Singh Koura and Marie-Anne Lachaux and Thibaut Lavril and Jenya Lee and Diana Liskovich and Yinghai Lu and Yuning Mao and Xavier Martinet and Todor Mihaylov and Pushkar Mishra and Igor Molybog and Yixin Nie and Andrew Poulton and Jeremy Reizenstein and Rashi Rungta and Kalyan Saladi and Alan Schelten and Ruan Silva and Eric Michael Smith and Ranjan Subramanian and Xiaoqing Ellen Tan and Binh Tang and Ross Taylor and Adina Williams and Jian Xiang Kuan and Puxin Xu and Zheng Yan and Iliyan Zarov and Yuchen Zhang and Angela Fan and Melanie Kambadur and Sharan Narang and Aurelien Rodriguez and Robert Stojnic and Sergey Edunov and Thomas Scialom},
      year={2023},
      eprint={2307.09288},
      archivePrefix={arXiv},
      primaryClass={cs.CL}
}

@ARTICLE{LoRA,
      title={LoRA: Low-Rank Adaptation of Large Language Models}, 
      author={Edward J. Hu and Yelong Shen and Phillip Wallis and Zeyuan Allen-Zhu and Yuanzhi Li and Shean Wang and Lu Wang and Weizhu Chen},
      year={2021},
      eprint={2106.09685},
      archivePrefix={arXiv},
      primaryClass={cs.CL}
}

@ARTICLE{MultimodalTransformerFusion,
       author = {{Xu}, Peng and {Zhu}, Xiatian and {Clifton}, David A.},
        title = "{Multimodal Learning with Transformers: A Survey}",
      journal = {arXiv e-prints},
     keywords = {Computer Science - Computer Vision and Pattern Recognition, Computer Science - Machine Learning},
         year = 2022,
        month = jun,
          %eid = {arXiv:2206.06488},
        %pages = {arXiv:2206.06488},
%archivePrefix = {arXiv},
       eprint = {2206.06488},
 %primaryClass = {cs.CV},
       %adsurl = {https://ui.adsabs.harvard.edu/abs/2022arXiv220606488X},
      adsnote = {Provided by the SAO/NASA Astrophysics Data System}
}

@inproceedings{16x16WORDS,
	title={An Image is Worth 16x16 Words: Transformers for Image Recognition at Scale},
	author={Alexey Dosovitskiy and Lucas Beyer and Alexander Kolesnikov and Dirk Weissenborn and Xiaohua Zhai and Thomas Unterthiner and Mostafa Dehghani and Matthias Minderer and Georg Heigold and Sylvain Gelly and Jakob Uszkoreit and Neil Houlsby},
	booktitle={International Conference on Learning Representations},
	year={2021}
	%url={https://openreview.net/forum?id=YicbFdNTTy}
}

@inproceedings{AttentionIsAll,
	author = {Vaswani, Ashish and Shazeer, Noam and Parmar, Niki and Uszkoreit, Jakob and Jones, Llion and Gomez, Aidan N and Kaiser, \L ukasz and Polosukhin, Illia},
	booktitle = {Advances in Neural Information Processing Systems},
	editor = {I. Guyon and U. Von Luxburg and S. Bengio and H. Wallach and R. Fergus and S. Vishwanathan and R. Garnett},
	pages = {},
	publisher = {Curran Associates, Inc.},
	title = {Attention is All you Need},
	%url = {https://proceedings.neurips.cc/paper/2017/file/3f5ee243547dee91fbd053c1c4a845aa-Paper.pdf},
	volume = {30},
	year = {2017}
}

@misc{OpenI_dataset,
	title = {Open-i: An open access biomedical search engine},
	howpublished = {\url{https://openi.nlm.nih.gov}},
	note = {Accessed: 2023-12-20}
}

@ARTICLE{OpenI_dataset_creator,
	title    = "Preparing a collection of radiology examinations for distribution
	and retrieval",
	author   = "Demner-Fushman, Dina and Kohli, Marc D and Rosenman, Marc B and
	Shooshan, Sonya E and Rodriguez, Laritza and Antani, Sameer and
	Thoma, George R and McDonald, Clement J",
	journal  = "J Am Med Inform Assoc",
	volume   =  23,
	number   =  2,
	pages    = "304--310",
	month    =  jul,
	year     =  2015,
	address  = "England",
	language = "en"
}

@inproceedings{BERT,
	title={BERT: Pre-training of Deep Bidirectional Transformers for Language Understanding},
	author={Jacob Devlin and Ming-Wei Chang and Kenton Lee and Kristina Toutanova},
	booktitle={North American Chapter of the Association for Computational Linguistics},
	year={2019},
	%url={https://api.semanticscholar.org/CorpusID:52967399}
}

@article{PEFT,
	title =        {PEFT: State-of-the-art Parameter-Efficient Fine-Tuning methods},
	author =       {Sourab Mangrulkar and Sylvain Gugger and Lysandre Debut and Younes Belkada and Sayak Paul and Benjamin Bossan},
	%url = {\url{https://github.com/huggingface/peft}},
	year =         {2022}
}

@inproceedings{PYTORCH,
	author = {Paszke, Adam and Gross, Sam and Massa, Francisco and Lerer, Adam and Bradbury, James and Chanan, Gregory and Killeen, Trevor and Lin, Zeming and Gimelshein, Natalia and Antiga, Luca and Desmaison, Alban and Kopf, Andreas and Yang, Edward and DeVito, Zachary and Raison, Martin and Tejani, Alykhan and Chilamkurthy, Sasank and Steiner, Benoit and Fang, Lu and Bai, Junjie and Chintala, Soumith},
	booktitle = {Advances in Neural Information Processing Systems},
	editor = {H. Wallach and H. Larochelle and A. Beygelzimer and F. d\textquotesingle Alch\'{e}-Buc and E. Fox and R. Garnett},
	pages = {},
	publisher = {Curran Associates, Inc.},
	title = {PyTorch: An Imperative Style, High-Performance Deep Learning Library},
	%url = {https://proceedings.neurips.cc/paper/2019/file/bdbca288fee7f92f2bfa9f7012727740-Paper.pdf},
	volume = {32},
	year = {2019}
}

@article{Dropout,
	author = {Srivastava, Nitish and Hinton, Geoffrey and Krizhevsky, Alex and Sutskever, Ilya and Salakhutdinov, Ruslan},
	year = {2014},
	month = {06},
	pages = {1929-1958},
	title = {Dropout: A Simple Way to Prevent Neural Networks from Overfitting},
	volume = {15},
	journal = {Journal of Machine Learning Research}
}

@misc{DeutscherRundfunk23,
title = {KI in der Klinik: Wie weit sind wir in Deutschland?; Interview Prof. Kai Wehkamp},
howpublished = {\url{https://www.deutschlandfunk.de/ki-in-der-klinik-wie-weit-sind-wir-in-deutschland-interview-prof-kai-wehkamp-dlf-8e2edd76-100.html}},
year = 2023,
month = 05,
day = 09,
note = {Accessed: 2024-31-01},
}
\end{document}